\documentclass{article}

\PassOptionsToPackage{numbers}{natbib}
\usepackage[main, final]{neurips_2026}

\usepackage[utf8]{inputenc} % allow utf-8 input
\usepackage[T1]{fontenc}    % use 8-bit T1 fonts
\usepackage{hyperref}       % hyperlinks
\usepackage{url}            % simple URL typesetting
\usepackage{booktabs}       % professional-quality tables
\usepackage{amsfonts}       % blackboard math symbols
\usepackage{nicefrac}       % compact symbols for 1/2, etc.
\usepackage{microtype}      % microtypography
\usepackage{xcolor}         % colors

\title{BayesRAG: Probabilistic Mutual Evidence Corroboration for Multimodal \\ Retrieval-Augmented Generation}

\author{
  \textbf{Xuan Li\textsuperscript{1}\thanks{Equal contribution.}} \quad
  \textbf{Junnan Zhu\textsuperscript{2}\footnotemark[1]} \quad
  \textbf{Yining Wang\textsuperscript{3}} \quad
  \textbf{Haocai Luo\textsuperscript{1}} \\
  \textbf{Shengping Liu\textsuperscript{3}} \quad
  \textbf{Jiaen Liang\textsuperscript{3}} \quad
  \textbf{Ying Fu\textsuperscript{3}} \quad
  \textbf{Wei Huang\textsuperscript{3}} \quad
  \textbf{Jun Yu\textsuperscript{1}\thanks{Corresponding author: \texttt{harryjun@ustc.edu.cn}.}}
  \\
  \textsuperscript{1}University of Science and Technology of China, % Department of Automation,
  \\
  \textsuperscript{2}MAIS, Institute of Automation, Chinese Academy of Sciences
  \\
  \textsuperscript{3}Unisound AI Technology Co., Ltd,
}

\usepackage{personal_preamble}
\usepackage{placeins} % Keep the ablation table after the wrapped recall table.

\begin{document}
\maketitle
\setcounter{footnote}{0} % Start numeric body footnotes independently of author notes.

\begin{abstract}
  Retrieval-Augmented Generation (RAG) has become a pivotal paradigm for Large Language Models (LLMs), yet current approaches struggle with visually rich documents by treating text and images as isolated retrieval targets. Existing methods relying solely on cosine similarity often fail to capture the semantic reinforcement provided by cross-modal alignment and layout-induced coherence.
  To address these limitations, we propose \textbf{BayesRAG}, a novel multimodal retrieval framework grounded in Bayesian inference and Dempster-Shafer evidence theory. Unlike traditional approaches that rank candidates strictly by similarity, BayesRAG models the intrinsic consistency of retrieved candidates across modalities as probabilistic evidence to refine retrieval confidence.
  Specifically, our method computes the posterior association probability for combinations of multimodal retrieval results, prioritizing text-image pairs that mutually corroborate each other in terms of both semantics and layout. Extensive experiments demonstrate that BayesRAG significantly outperforms state-of-the-art (SOTA) methods on challenging multimodal benchmarks. This study establishes a new paradigm for multimodal retrieval fusion that effectively resolves the isolation of heterogeneous modalities through an evidence fusion mechanism and enhances the robustness of retrieval outcomes. Our code is available at \url{https://github.com/TioeAre/BayesRAG}.
\end{abstract}

\section{Introduction}
\label{sec:introduction}

RAG has fundamentally enhanced the generalization capabilities of LLMs by grounding them in external domain knowledge. However, real-world knowledge, ranging from academic papers to industrial reports, is predominantly presented in visually rich documents. In these documents, visual elements such as charts and diagrams are not merely decorative add-ons, but are intrinsically intertwined with textual explanations to convey complex semantics. Consequently, relying solely on unimodal text retrieval is insufficient. Verifying the semantic consistency between visual and textual evidence is paramount for building robust RAG systems.

In practice, RAG systems navigate vast repositories containing thousands of pages. While the advent of vision-language models has enabled multimodal retrieval, current SOTA approaches primarily focus on improving recall by expanding the Top-$k$ window and performing a naive union of text and image retrieval results. We argue that this ``bag-of-evidence'' approach is fundamentally flawed. By treating modalities as independent channels and simply concatenating their representations, existing methods fail to capture the critical \textit{semantic interaction} between modalities. Specifically, they cannot distinguish whether a retrieved image and a text chunk are logically corroborated or merely statistically similar (e.g., sharing common keywords but divergent contexts). This inability leads to a collection of high-similarity yet semantically disconnected fragments, which introduces noise and confounds the downstream answer generator.

To address modal disparity, some approaches employ Image-to-Text transduction or Knowledge Graphs (KGs) to unify information into a textual modality. However, these methods suffer from severe information and structural loss. Critical insights in knowledge-intensive domains are often encoded in non-textual formats that are ``ineffable'', which is difficult to translate losslessly into descriptions~\citep{lumer2025comparison}. Furthermore, these linearization processes remove document layout information~\citep{xiang2025use}. We consider that document layout is not merely a stylistic choice but serves as an implicit logical navigation map designed by the author. Standard chunking strategies disrupt this topological structure, severing the logical ties between spatially adjacent evidence.

\begin{figure*}[tp]
  \centering
  \includegraphics[width=\textwidth]{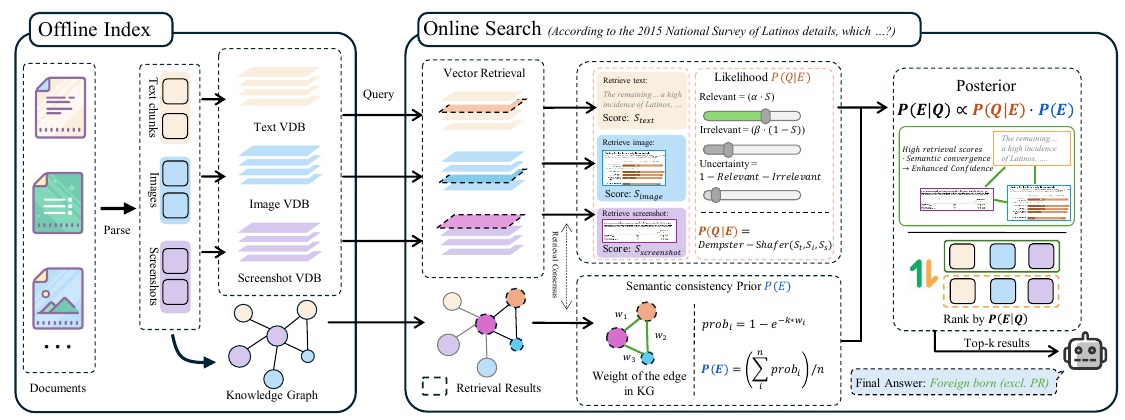}
  \caption[Overview]
  {
    \textmd{The architecture of BayesRAG. We reconceptualize multimodal retrieval as a Bayesian inference process. Our method computes the semantic likelihood score $P(Q|E)$ and derives the consistency prior $P(E)$ via knowledge graph topology. The final evidence tuples are re-ranked based on the posterior evidence score $P(E|Q)$, favoring semantic relevance, mutual corroboration, and logical self-consistency.}
  }
  \label{fig:framework}
  \vspace{-1.0em}
\end{figure*}

In summary, developing a trustworthy multimodal RAG system necessitates addressing three pivotal challenges: (1) \textbf{Preserving Multimodal Fidelity:} The system should move beyond lossy translation to perform native representation learning~\citep{mei2025surveymultimodalretrievalaugmentedgeneration}. (2) \textbf{Cross-Modal Semantic Corroboration:} Achieving high confidence requires a mechanism for evidence verification. The system needs to facilitate interaction between retrieved modalities, identifying instances where visual and textual candidates semantically align. Such alignment should be treated as reinforcing evidence, thereby assigning higher posterior confidence to mutually corroborating retrieval results. (3) \textbf{Layout-Induced Structural Understanding:} Beyond content semantics, the system should be able to recover the contextual relationships of retrieved elements within the original document, which allows for decoding the implicit semantic topology within the document layout~\citep{tong2025hkragholisticknowledgeretrievalaugmented}, thereby restoring connections that are frequently severed by traditional chunking strategies.

To address these challenges, we propose \textbf{BayesRAG}. Diverging from traditional approaches that rely solely on unimodal similarity matching, we reconceptualize multimodal retrieval as a process of \textit{probabilistic evidence fusion}. Our core intuition addresses the prevalent issue of semantic inconsistency in heterogeneous retrieval; this is exemplified by scenarios where text retrieval suggests the fruit ``apple'' while visual retrieval identifies the corporate logo of ``Apple Inc.'' Under such circumstances, independent similarity scores remain high for both yet fail to capture the fatal discordance between them. BayesRAG leverages Bayesian inference to elegantly resolve this dilemma by utilizing similarity scores from embedding models as the \textit{likelihood} and modeling cross-modal semantic consistency as the \textit{Prior Belief}. This formulation mathematically embodies the intuition that \textit{``only text-image pairs that are semantically consistent deserve higher trust,''} and it thereby dynamically penalizes conflicting evidence during the fusion.

By computing the posterior evidence score, BayesRAG automatically down-weights retrieval candidates that exhibit high unimodal scores yet contradict each other. This mechanism effectively isolates high-fidelity evidence combinations that are mutually corroborating, simultaneously preserving native multimodal information and ensuring the logical rigor of cross-modal reasoning. Our main contributions are summarized as follows:

\begin{itemize}[leftmargin=10pt]
  \item We establish a unified probabilistic framework that reconceptualizes multimodal retrieval as an evidence fusion process; this paradigm effectively preserves the native layout structure often lost in traditional strategies.

  \item We introduce a Bayesian inference mechanism grounded in Dempster-Shafer theory to model semantic consistency; this approach resolves cross-modal conflicts by prioritizing mutually corroborating evidence.

  \item Extensive experiments demonstrate that BayesRAG achieves SOTA performance on challenging benchmarks; our analysis confirms that modeling evidence consistency effectively eliminates modal conflicts and semantic inconsistencies prevalent in existing baselines.
\end{itemize}

\section{Method}
\label{sec:method}

In this section, we present \textbf{BayesRAG}, a probabilistic framework for multimodal retrieval.

\subsection{Retrieval as Evidence Fusion}
\label{subsec:method_overview}

Traditional multimodal RAG systems typically treat retrieval as independent feature-matching tasks, where text and images are ranked solely by their semantic similarity to the query. This often leads to modal conflict, where a high-scoring text and a high-scoring image may contradict each other or originate from unrelated contexts.

BayesRAG redefines retrieval as a \textbf{Probabilistic Evidence Fusion} process. We aim to identify an evidence tuple $E = (t,v,s)$, containing a text chunk, a visual crop, and a full-page screenshot, that is not only relevant to the query but also \textit{internally consistent.} Our intuition is modeled via Bayesian inference: \textbf{Prior} (Internal Consistency): Before observing the query, how likely is it that this text and image are related? This is determined by the inherent consistency of the tuple $E$, exemplified by factors such as semantic alignment or narrative proximity in the source document. \textbf{Likelihood} (Query Support): How strongly do the retrieved modalities support the query? We estimate this support using vector similarities combined through belief functions. \textbf{Posterior} (Evidence Ranking): The updated evidence score used to rank the tuple for the query.

\subsection{Problem Formulation}
\label{sec:problem_formulation}

Let $\mathcal{D}$ be a multimodal document corpus. We define the sample space $\mathcal{E}$ as the set of all possible multimodal evidence tuples. A single evidence candidate is denoted as $E = (t, v, s) \in \mathcal{E}$, where $t$ is one retrieved text chunk, $v$ is one locally cropped visual element (e.g., a figure, chart, or table region), and $s$ is one full-page PDF screenshot. The crop preserves localized visual detail, whereas the screenshot retains surrounding content and page-level layout.

The three streams are retrieved independently, and their candidates are combined into tuples at query time. A component can participate in several tuples, allowing its relevance to be assessed alongside different cross-modal partners.

Let $Q$ denote the observed user query. Our goal is to rank candidate tuples by their query-conditioned evidence score $P(E|Q)$. We express this ranking objective through Bayes' theorem:
\begin{equation}
  P(E|Q) = \frac{P(Q|E) \cdot P(E)}{P(Q)}
\end{equation}
Since $P(Q)$ (the marginal likelihood of the query) is constant for all candidates during the ranking phase, we simplify the objective function to:
\begin{equation}
  P(E|Q) \propto \underbrace{P(Q|E)}_{\text{Embedding Likelihood}} \cdot \underbrace{P(E)}_{\text{Consistency Prior}}
  \label{eq:objective}
  %   \small
\end{equation}

The two factors in Equation~\ref{eq:objective} capture complementary aspects of evidence selection. The likelihood measures support for the query, while the prior describes relationships among the retrieved items in the source document. A figure and its explanation can therefore support one another even when their individual query similarities differ. Their combined score favors candidates that match the query and form a coherent evidence set.

\subsection{Bayesian Evidence Modeling}
\label{subsec:bayes_model}

\textbf{Modeling Likelihood with Belief Functions.}
The likelihood $P(Q|E)$ estimates the query-conditioned support from the evidence $E$. To robustly handle the uncertainty in embedding cosine similarity scores (e.g., distinguishing between ``irrelevant'' and ``uncertain''), we incorporate Dempster-Shafer evidence theory~\citep{dempster1967upper,shafer1976mathematical}.

\begin{wraptable}{r}{0.48\textwidth}
  \vspace{-1em}
  \centering
  \scriptsize
  \setlength{\tabcolsep}{3pt}
  \renewcommand{\arraystretch}{0.95}
  \caption{Mass functions from cosine similarity scores. $\alpha, \beta \in [0,1]$ denote modality trust.}
  \begin{tabular*}{0.9\linewidth}{@{\extracolsep{\fill}}ll@{}}
    \toprule
    \textbf{Hypothesis} & \textbf{Formulation} \\ \midrule
    Evidence ($e$) & $m(\text{Y}) = \alpha S_i$ \\
    Not Evidence ($\bar{e}$) & $m(\text{N}) = \beta (1 - S_i)$ \\
    Uncertainty ($\Theta$) & $m(\Theta) = 1 - m(\text{Y}) - m(\text{N})$ \\
    \bottomrule
  \end{tabular*}
  \label{tab:mass_function}
\end{wraptable}

Since retrieval scores from different embedding models (e.g., dense text embeddings vs. visual embeddings) often inhabit different numerical ranges, direct combination is infeasible. We first normalize raw scores into the unit interval $[0, 1]$. For each modality $i \in \{t, v, s\}$, let $S_i \in [0,1]$ be the normalized cosine similarity score with the query. We construct a mass function $m_i$ over the frame of discernment $\Theta = \{\text{Y}, \text{N}\}$ (Relevant, Irrelevant), as shown in Table~\ref{tab:mass_function}.

We fuse the three modalities by applying Dempster's Rule sequentially: we first merge two sources (e.g., text and vision) into an intermediate belief state, and then merge that result with the third source (screenshot). In this sense, the operator $\oplus$ in $m_{comb} = m_i \oplus m_j$ can be read as a \emph{conflict-aware, normalized merge} between two belief distributions rather than as ordinary addition. Concretely, when combining a current belief $m_{curr}$ with a new source $m_{new}$, we first compute the conflict coefficient
\begin{equation}
  K = m_{curr}(\text{Y}) m_{new}(\text{N}) + m_{curr}(\text{N}) m_{new}(\text{Y}),
\end{equation}
which measures how much one source supports relevance while the other supports irrelevance. When $K$ exceeds a conflict threshold, we regard the two sources as fundamentally contradictory and set the likelihood to $0$. Otherwise, we merge the non-conflicting evidence and renormalize it by $1-K$:
\begin{equation}
  m_{comb}(H) = \frac{\sum_{A \cap B = H} m_{curr}(A) m_{new}(B)}{1 - K}, \quad \varnothing\ne H\subseteq\Theta,\ A,B\subseteq\Theta.
\end{equation}
The sum includes hypothesis sets whose intersection is $H$. For the relevant hypothesis, this merge becomes
\begin{equation}
  m_{comb}(\text{Y}) = \frac{m_{curr}(\text{Y}) m_{new}(\text{Y}) + m_{curr}(\text{Y}) m_{new}(\Theta) + m_{curr}(\Theta) m_{new}(\text{Y})}{1 - K}.
\end{equation}
The numerator separates two forms of compatible support. Its first term combines sources that both assign mass to relevance. The other terms preserve support from one source when the other retains uncommitted mass. For example, a supporting text chunk can be combined with an uncertain page screenshot through these terms. Opposing relevance and irrelevance assignments contribute to $K$. The merge thus distinguishes agreement, incomplete support, and disagreement.

The likelihood is then projected from the combined belief function:
\begin{equation}
  P(Q|E) \approx \text{BetP}_{joint}(\text{Y}) = m_{final}(Y) + \frac{m_{final}(\Theta)}{2}
\end{equation}
The pignistic transformation assigns half of the remaining uncertainty mass to relevance and half to irrelevance. This produces a single ranking score while retaining the distinction between direct support and uncommitted belief during fusion (Appendix~\ref{app:likelihood_estimation}).

\paragraph{Modeling the Consistency Prior $P(E)$.}
The prior $P(E)$ scores whether the tuple $E=(t, v, s)$ constitutes a meaningful, coherent unit of information \textit{independent} of the user's query. This captures the ``topological'' or ``physical'' reality of the document structure. If the text $t$ and image $v$ are semantically unrelated or spatially distant in the source document, $P(E)$ should be low. We model this as:
\begin{equation}
  P(E) = f_{\text{consistency}}(t, v, s)
\end{equation}
where $f_{\text{consistency}}$ maps either geometric proximity or semantic connectivity into a prior score. Concretely, we instantiate $P(E)$ in two ways. For the \textbf{Layout Prior}, we assign a high prior when the text, image, and screenshot come from the same document, lie on the same or adjacent pages, and the text--image distance is below a threshold relative to the page diagonal; otherwise we assign a small fallback value:
\begin{equation}
  P(E)_{\text{layout}} =
  \begin{cases}
    1,        & \text{if } \mathcal{C}_{\text{dist}} \land \mathcal{C}_{\text{page}}, \\
    \epsilon, & \text{otherwise.}
  \end{cases}
\end{equation}
Here, distance is measured between the centers of the MinerU bounding boxes. On a page of width $W$ and height $H$, the normalized distance is $\|c_t-c_v\|_2/\sqrt{W^2+H^2}$. Adjacent pages are placed on a shared canvas before computing the distance (Appendix~\ref{app:layout_prior}).

For the \textbf{Graph-Topology Prior}, each retrieved text chunk, visual region, or page is represented through its corresponding graph node. Extracted entities are mapped back to their source regions, and graph relations connect the associated nodes. For a pair of nodes $(u,v)$ in the graph, $S_{uv}=\sum_{r\in\mathcal{R}_{u,v}}w_r$ aggregates the weights of relations connecting them. We convert this strength into $p_{uv}=P_{\text{edge}}(u,v)=1-\exp(-\kappa S_{uv})$, and score the evidence tuple by averaging its three connections:
\begin{equation}
  P(E)_{\text{graph}} = \frac{p_{tv} + p_{ts} + p_{vs}}{3}.
\end{equation}
Here, $p_{tv}$ links the text chunk to the visual crop, while $p_{ts}$ and $p_{vs}$ connect these local items to the page screenshot. A tuple supported by several strong connections receives a higher average than one supported by a single connection of the same strength. This combines local cross-modal linkage with connections to page-level context.

The full model uses the Graph-Topology Prior; the Layout Prior provides an alternative based on geometric proximity. Section~\ref{sec:ablation_experiments} compares these structural cues.

\subsection{Evidence Selection and Context Construction}
\label{sec:evidence_selection}

Given a query, BayesRAG retrieves an initial pool of 512 candidates from each modality. We retain high-scoring embedding matches and construct additional tuples from compatible candidates. Source-document compatibility and graph connections restrict which text chunks, visual crops, and screenshots are combined. This sparse construction allows cross-modal support to influence the selection of candidates below the leading embedding matches.

We score each retained tuple by the product in Equation~\ref{eq:objective} and rank the tuples by this posterior evidence score. The top 20 items from each modality form the generator input, preserving text chunks, cropped images, and full-page screenshots in the context. Appendix~\ref{app:tuple_construction} describes the candidate-construction procedure.

\section{Experiments}
\label{sec:experiment}

\subsection{Experimental Settings}

\textbf{Evaluation Benchmarks}. We evaluate BayesRAG on two challenging multimodal Document QA benchmarks: DocBench \citep{zou2025docbench} and MMLongBench-Doc \citep{ma2024mmlongbench}. DocBench comprises 229 documents across five domains (e.g., Finance, Law) with 1,102 QA pairs, featuring substantial length (avg. 66 pages, $\sim$46k tokens) to test long-context capabilities. MMLongBench-Doc complements this with 135 documents spanning 7 diverse types and 1,091 questions\footnote{\url{https://huggingface.co/datasets/yubo2333/MMLongBench-Doc}}. Together, these datasets cover over 2,000 questions in varied real-world scenarios, providing a testbed for multimodal long-document understanding.

\paragraph{Baselines.} To evaluate the effectiveness of BayesRAG, we compare it against representative multimodal RAG systems, covering modular, visual-centric, agentic, and graph-based paradigms: \textbf{RAGFlow}\footnote{\url{https://github.com/infiniflow/ragflow}}: A widely adopted open-source RAG engine that integrates retrieval pipelines with agentic capabilities. It serves as a representative baseline for standard, modularized RAG implementations used in industry. \textbf{VisRAG} \citep{yu2025visrag}: A visual-centric framework that bypasses traditional OCR parsing. Instead of extracting text, VisRAG treats document pages directly as images, utilizing VLMs for both visual embedding and generation. \textbf{ViDoRAG} \citep{wang2025vidorag}: An agent-based approach that employs a Gaussian Mixture Model (GMM) for multimodal fusion. It introduces an iterative agentic workflow (consisting of Explore, Summarize, and Reflect stages) to verify the reliability of retrieved evidence dynamically. \textbf{RAGAnything} \citep{guo2025rag}: A knowledge-graph-based system that reconceptualizes multimodal content as interconnected entities. It employs a dual-graph construction mechanism and hybrid retrieval, combining structural navigation within the KG with semantic vector matching to locate cross-modal evidence. We also evaluate \textbf{Qwen3-VL-32B} directly on the target document without RAG.

\paragraph{Evaluation Metrics.} For \textbf{DocBench}, we strictly follow the official evaluation protocol, utilizing the standard LLM-based scoring mechanism to ensure fair comparison with existing baselines.
For \textbf{MMLongBench-Doc}, in addition to reporting the official rule-based accuracy (Acc), we introduce a supplementary LLM-based evaluation. This is designed to mitigate false negatives inherent in rigid string matching, specifically addressing cases of semantic equivalence (e.g., abbreviations or paraphrasing) that rule-based metrics fail to capture.

\paragraph{Experimental Setup.} We employ the Qwen3-VL-32B as the final response generator across experiments. We utilize MinerU~\citep{niu2025mineru25decoupledvisionlanguagemodel} for document parsing. We employ Qwen3-Embedding-4B~\citep{zhang2025qwen3} to encode textual chunks. For image elements, we utilize PE-Core-G14-448~\citep{bolya2025perception}. Additionally, we use colnomic-embed-multimodal-3b\footnote{\url{https://huggingface.co/nomic-ai/colnomic-embed-multimodal-3b}} to generate page-level embeddings for global layout understanding. Retrieved text candidates are refined using bge-reranker-v2-m3~\citep{multi2024chen}. For the construction of the Knowledge Graph, we leverage the Qwen3-VL-32B~\citep{bai2025qwen3} to extract entities and relations from both text and images. We index the document corpus in a shared vector database and knowledge graph. For each query, we use the benchmark-provided \texttt{doc\_id} to retrieve evidence from the corresponding source document. The main experiments use this single-document setting and results for retrieval across the full collection are reported in Appendix~\ref{app:generator_impact}.
We evaluate all methods with the same answer generator and source-document restriction. BayesRAG first retrieves the top 512 candidates from each modality. We then rerank the retrieved evidence using the Bayesian posterior scores and pass the top-20 items from each modality to the generator. Other retrieval methods select their top-20 evidence items using their respective retrieval and ranking procedures. Further implementation details and configurations for the earlier experiments are provided in Appendix~\ref{app:reproducibility}.

\subsection{Main Results}
\label{subsec:main_results}

\begin{wraptable}{r}{0.55\textwidth}
  \vspace{-1.0em}
  \centering
  \scriptsize
  \setlength{\tabcolsep}{3pt}
  \renewcommand{\arraystretch}{0.95}
  \caption{Accuracy (\%) with Qwen3-VL-32B. \textbf{Bold}: best; \underline{underline}: second best.}
  \resizebox{0.98\linewidth}{!}{%
    \begin{tabular}{@{}lcccccc@{}}
      \toprule
      \multirow{2}{*}{\textbf{Method}} & \multicolumn{5}{c}{\textbf{Domains}} & \multirow{2}{*}{\textbf{Overall}} \\ \cmidrule(lr){2-6}
      & Aca. & Fin. & Gov. & Law & News & \\ \midrule
      Qwen3-VL-32B & {\ul 76.6} & 45.8 & 66.2 & 60.2 & 80.8 & 65.0 \\
      VisRAG & 60.7 & 63.2 & 65.5 & 63.4 & \textbf{88.4} & 66.8 \\
      ViDoRAG & 61.7 & \textbf{76.4} & 64.2 & 70.2 & 84.3 & 70.9 \\
      RAGFlow & 64.7 & 70.5 & 66.9 & 62.8 & 37.8 & 62.0 \\
      RAGAnything & 71.0 & 68.1 & {\ul 73.0} & {\ul 73.8} & 80.2 & {\ul 72.4} \\
      \rowcolor{green!12} BayesRAG (ours) & \textbf{80.2} & {\ul 75.3} & \textbf{77.0} & \textbf{80.6} & {\ul 85.5} & \textbf{79.4} \\ \bottomrule
    \end{tabular}%
  }
  \label{tab:docbench_results}
\end{wraptable}

\paragraph{Performance comparison.} Table~\ref{tab:docbench_results} compares the methods on DocBench. BayesRAG achieves the highest overall accuracy of \textbf{79.4\%}, surpassing the strongest overall baseline, RAGAnything (72.4\%), by \textbf{7.0} percentage points. It outperforms RAGAnything across all five domains, with gains of 7.2 points in \textit{Finance} and 5.3 points in \textit{News}. Alongside query similarity, BayesRAG uses evidence consistency to favor text and visual elements that support one another. This gives the retrieval system a common way to select evidence across domains with different mixtures of narrative text, figures, and page layout.

As shown in Table~\ref{tab:mmlongbench_results}, BayesRAG also achieves the highest overall results on MMLongBench-Doc, with \textbf{57.1\%} accuracy and an LLM score of \textbf{62.3\%}. These exceed RAGAnything by \textbf{2.7} and \textbf{4.9} percentage points, respectively. The LLM score improves over RAGAnything in all seven domains. Notable gains include \textit{Guidebooks} (+3.9 points in accuracy over the runner-up) and \textit{Financial Reports} (+12.8 points in LLM score over RAGAnything). BayesRAG also ranks first in \textit{Academic} under both metrics. Such documents often combine textual descriptions, charts, and page-level context. By preserving these complementary views, BayesRAG can use their relationships during evidence selection instead of relying on an isolated text or image match. These gains highlight its strength in combining textual, visual, and layout evidence for multimodal document question answering.

% \usepackage{booktabs}
% \usepackage{multirow}
% \usepackage{graphicx}
% \usepackage[normalem]{ulem}
\begin{table*}[tp]
  \caption{Accuracy (\%) and LLM score (\%) on MMLongBench-Doc with Qwen3-VL-32B across different domains and overall performance. Best performance is highlighted in \textbf{bold} and second-best is \underline{underlined}. Domain categories include Research Reports/Introductions (Reports), Tutorials/Workshops (Tutorials), Academic Papers (Academic), Guidebooks, Brochures, Administration/Industry Files (Industry), and Financial Reports (Financial). Acc uses the benchmark verification procedure; Score uses LLM-based judging.}
  \resizebox{\textwidth}{!}{%
    \begin{tabular}{@{}lcccccccccccccccc@{}}
      \toprule
      \multirow{2}{*}{\textbf{Method}} &
      \multicolumn{2}{c}{\textbf{Reports}} &
      \multicolumn{2}{c}{\textbf{Tutorials}} &
      \multicolumn{2}{c}{\textbf{Academic}} &
      \multicolumn{2}{c}{\textbf{Guidebooks}} &
      \multicolumn{2}{c}{\textbf{Brochures}} &
      \multicolumn{2}{c}{\textbf{Industry}} &
      \multicolumn{2}{c}{\textbf{Financial}} &
      \multicolumn{2}{c}{\textbf{Overall}} \\
      \cmidrule(lr){2-3}\cmidrule(lr){4-5}\cmidrule(lr){6-7}\cmidrule(lr){8-9}\cmidrule(lr){10-11}\cmidrule(lr){12-13}\cmidrule(lr){14-15}\cmidrule(lr){16-17}
      & Acc  & Score & Acc  & Score & Acc  & Score & Acc  & Score & Acc  & Score & Acc           & Score         & Acc  & Score & Acc  & Score \\ \midrule
      Qwen3-VL-32B & 54.6 & 57.7 & \textbf{59.0} & \textbf{64.7} & 45.1 & 46.6 & 48.1 & 52.6 & {\ul 52.5} & \textbf{55.4} & \textbf{55.6} & \textbf{67.9} & 44.4 & 48.7 & 51.2 & 55.4 \\
      VisRAG & 45.7 & 48.5 & 50.4 & 51.8 & 32.4 & 34.8 & 46.2 & 48.1 & 42.6 & 44.6 & 45.7 & 50.6 & 45.3 & 44.4 & 43.5 & 45.6 \\
      ViDoRAG & 47.1 & 49.5 & 51.8 & 54.0 & 45.1 & 44.6 & 44.2 & 46.2 & 33.7 & 34.7 & 44.4 & 48.1 & 47.0 & 46.2 & 45.5 & 46.8 \\
      RAGFlow & 47.8 & 51.9 & 49.6 & 52.5 & 45.6 & 46.6 & 44.9 & 48.1 & 38.6 & 40.6 & 35.8 & 37.0 & 45.3 & 45.3 & 45.2 & 47.6 \\
      RAGAnything & {\ul 58.7} & {\ul 61.4} & 55.4 & 60.4 & {\ul 53.4} & {\ul 54.4} & {\ul 53.2} & {\ul 55.8} & 47.5 & {\ul 49.5} & {\ul 51.9} & 60.5 & {\ul 53.0} & {\ul 55.6} & {\ul 54.4} & {\ul 57.4} \\
      \rowcolor{green!12} BayesRAG (ours) & \textbf{59.0} & \textbf{64.2} & {\ul 57.6} & {\ul 64.0} & \textbf{56.4} & \textbf{59.3} & \textbf{57.1} & \textbf{61.5} & \textbf{54.5} & \textbf{55.4} & 50.6 & {\ul 61.7} & \textbf{59.8} & \textbf{68.4} & \textbf{57.1} & \textbf{62.3} \\ \bottomrule
    \end{tabular}%
  }
  \label{tab:mmlongbench_results}
  \vspace{-1em}
\end{table*}

% \usepackage{booktabs}
% \usepackage{graphicx}
\begin{wraptable}{r}{0.60\textwidth}
  \vspace{0.5em}
  \centering
  \scriptsize
  \setlength{\tabcolsep}{3pt}
  \renewcommand{\arraystretch}{0.95}
  \caption{Retrieval performance on MMLongBench-Doc.}
  \resizebox{0.98\linewidth}{!}{%
    \begin{tabular}{@{}lccccc@{}}
      \toprule
      \textbf{Method} & \textbf{Recall@1} & \textbf{Recall@3} & \textbf{Recall@5} & \textbf{Recall@10} & \textbf{Recall@20} \\ \midrule
      Vector Retrieval & 24.4 & 41.7 & 46.4 & 51.6 & 56.6 \\
      BayesRAG & 27.5 & 44.3 & 54.7 & 61.7 & 76.6 \\ \bottomrule
    \end{tabular}%
  }
  \label{tab:retrieval_recall}
  % \vspace{-0.5em}
\end{wraptable}

\noindent\textbf{Retrieval performance.} We compare BayesRAG against Vector Retrieval, which aggregates cosine similarity scores from text, image, and screenshot embeddings with a text reranker. Using the page-level evidence annotations in MMLongBench-Doc, we approximate retrieval recall before answer generation. Table~\ref{tab:retrieval_recall} shows improvements at every reported cutoff. Recall@1 increases from 24.4\% to 27.5\%, a gain of \textbf{3.1} points. Recall@20 rises from 56.6\% to 76.6\%, a gain of \textbf{20.0} points. These gains show improved recovery of annotated evidence both in the leading result and within the larger retrieved set. Consistency among text, crops, and screenshots provides an additional ranking signal beyond the independent query similarities.

\begin{wraptable}{r}{0.48\textwidth}
  \centering
  \caption{Modality ablation on MMLongBench-Doc. Recall@20 is reported in \%.}
  \label{tab:modality_ablation}
  \scriptsize
  \setlength{\tabcolsep}{3pt}
  \renewcommand{\arraystretch}{0.95}
  \begin{tabular*}{\linewidth}{@{\extracolsep{\fill}}lc@{}}
    \toprule
    \textbf{Evidence modalities} & \textbf{Recall@20 $\uparrow$} \\ \midrule
    Text + Image & 66.3 \\
    Text + Screenshot & 62.6 \\
    Image + Screenshot & 60.5 \\
    Text + Image + Screenshot & \textbf{76.6} \\
    \bottomrule
  \end{tabular*}
\end{wraptable}

\noindent\textbf{Complementarity of Evidence Modalities.} Table~\ref{tab:modality_ablation} compares the full three-view configuration with variants that omit one modality on MMLongBench-Doc. The full configuration achieves \textbf{76.6\%} Recall@20. Removing screenshots reduces recall by \textbf{10.3} percentage points, while removing images or text reduces it by 14.0 and 16.1 points, respectively. The largest reduction occurs without text, but both visual branches also contribute to the best retrieval result. In particular, screenshots remain useful alongside text chunks and cropped images despite their partial content overlap. These comparisons support retaining all three views for evidence selection.

\subsection{Ablation Experiments}
\label{sec:ablation_experiments}

Table~\ref{tab:ablation_study} compares changes to the likelihood model and consistency prior on MMLongBench-Doc, alongside the Embedding-RAG baseline. All variants use Qwen3-VL-32B for answer generation.

\noindent\textbf{Bridging the Visual Reranking Gap.} Embedding-RAG refines textual candidates with a text reranker, while visual evidence is selected using embedding similarities. BayesRAG applies posterior reranking to the broader candidate pool described in Section~\ref{sec:evidence_selection}. As shown in Table~\ref{tab:ablation_study}, BayesRAG (Full) scores 62.3\%, compared with 58.7\% for Embedding-RAG, a gain of 3.6 percentage points. Its ranking score combines query support with relationships among the retrieved items, allowing visual candidates to be assessed alongside textual and page-level evidence.

\noindent\textbf{Comparison of fusion strategy.} We compare Dempster-Shafer fusion with \textit{Linear Fusion}, which averages normalized embedding scores; both variants use the Graph prior. Linear Fusion achieves 60.0\%, whereas BayesRAG (Full) reaches 62.3\%, an improvement of 2.3 percentage points. Dempster-Shafer fusion leads in six of the seven domains. The largest gains occur in \textit{Brochures}, \textit{Guidebooks}, and \textit{Tutorials}, at 5.9, 4.4, and 4.3 points, respectively. These results favor combining relevance support through belief functions, which distinguish compatible support, uncertainty, and opposing signals during fusion. The fused scores then guide the ranking of evidence tuples.

\noindent\textbf{Spatial vs. Semantic Priors.} With Dempster-Shafer fusion fixed, the Layout Prior variant achieves 60.7\%, while the Graph-Topology Prior reaches 62.3\%, a gain of 1.6 percentage points. Graph topology leads in five of the seven domains, with the largest gains in \textit{Financial} and \textit{Tutorials}, at 5.2 and 5.0 points. Layout remains effective in \textit{Academic} and \textit{Industry}, where it achieves the highest scores among these variants. The comparison favors graph relationships for the overall benchmark while showing the value of geometric proximity in particular domains.

% \usepackage{booktabs}
% \usepackage{multirow}
% \usepackage{graphicx}
\begin{table*}
  \caption{Ablation study on MMLongBench-Doc with Qwen3-VL-32B (LLM score, \%). Variants change the likelihood model or consistency prior; Embedding-RAG uses independent retrieval and a text reranker. Bold marks the best result in each column. Parenthesized values give percentage-point differences from BayesRAG (Full).}
  \resizebox{\textwidth}{!}{%
    \begin{tabular}{@{}lccccccccccc@{}}
      \toprule
      \textbf{Method} &
      \textbf{
        \begin{tabular}[c]{@{}c@{}}Bayesian\\ Inference
      \end{tabular}} &
      \textbf{Likelihood} &
      \textbf{Prior} &
      \textbf{Reports} &
      \textbf{Tutorials} &
      \textbf{Academic} &
      \textbf{Guidebooks} &
      \textbf{Brochures} &
      \textbf{Industry} &
      \textbf{Financial} &
      \textbf{Overall} \\ \midrule
      BayesRAG (Full) & \Checkmark\xspace & Dempster-Shafer & \multicolumn{1}{c|}{Graph} & \textbf{64.2} & \textbf{64.0} & 59.3 & \textbf{61.5} & \textbf{55.4} & 61.7 & \textbf{68.4} & \textbf{62.3} \\
      BayesRAG (Linear Fusion) & \Checkmark\xspace & Linear Weighted & \multicolumn{1}{c|}{Graph} & 62.8 {\scriptsize($\downarrow$1.4)} & 59.7 {\scriptsize($\downarrow$4.3)} & 58.8 {\scriptsize($\downarrow$0.5)} & 57.1 {\scriptsize($\downarrow$4.4)} & 49.5 {\scriptsize($\downarrow$5.9)} & 65.4 {\scriptsize($\uparrow$3.7)} & 65.0 {\scriptsize($\downarrow$3.4)} & 60.0 {\scriptsize($\downarrow$2.3)} \\
      BayesRAG (Layout Prior) & \Checkmark\xspace & Dempster-Shafer & \multicolumn{1}{c|}{Layout} & 61.1 {\scriptsize($\downarrow$3.1)} & 59.0 {\scriptsize($\downarrow$5.0)} & \textbf{60.3} {\scriptsize($\uparrow$1.0)} & 60.9 {\scriptsize($\downarrow$0.6)} & 53.5 {\scriptsize($\downarrow$1.9)} & \textbf{67.9} {\scriptsize($\uparrow$6.2)} & 63.2 {\scriptsize($\downarrow$5.2)} & 60.7 {\scriptsize($\downarrow$1.6)} \\
      Embedding-RAG & \XSolidBrush\xspace & - & \multicolumn{1}{c|}{-} & 62.5 {\scriptsize($\downarrow$1.7)} & 58.3 {\scriptsize($\downarrow$5.7)} & 59.3 {\scriptsize(0.0)} & 53.8 {\scriptsize($\downarrow$7.7)} & 54.5 {\scriptsize($\downarrow$0.9)} & 60.5 {\scriptsize($\downarrow$1.2)} & 57.3 {\scriptsize($\downarrow$11.1)} & 58.7 {\scriptsize($\downarrow$3.6)} \\ \bottomrule
    \end{tabular}%
  }
  \label{tab:ablation_study}
  % \vspace{-5mm}
  % \vspace{-1.0em}
\end{table*}

\subsection{Hyperparameter Sensitivity}
\label{sec:hyperparameter_sensitivity}

\begin{wrapfigure}{r}{0.4\textwidth}
  \vspace{-1.5em}
  \centering
  \includegraphics[width=0.98\linewidth]{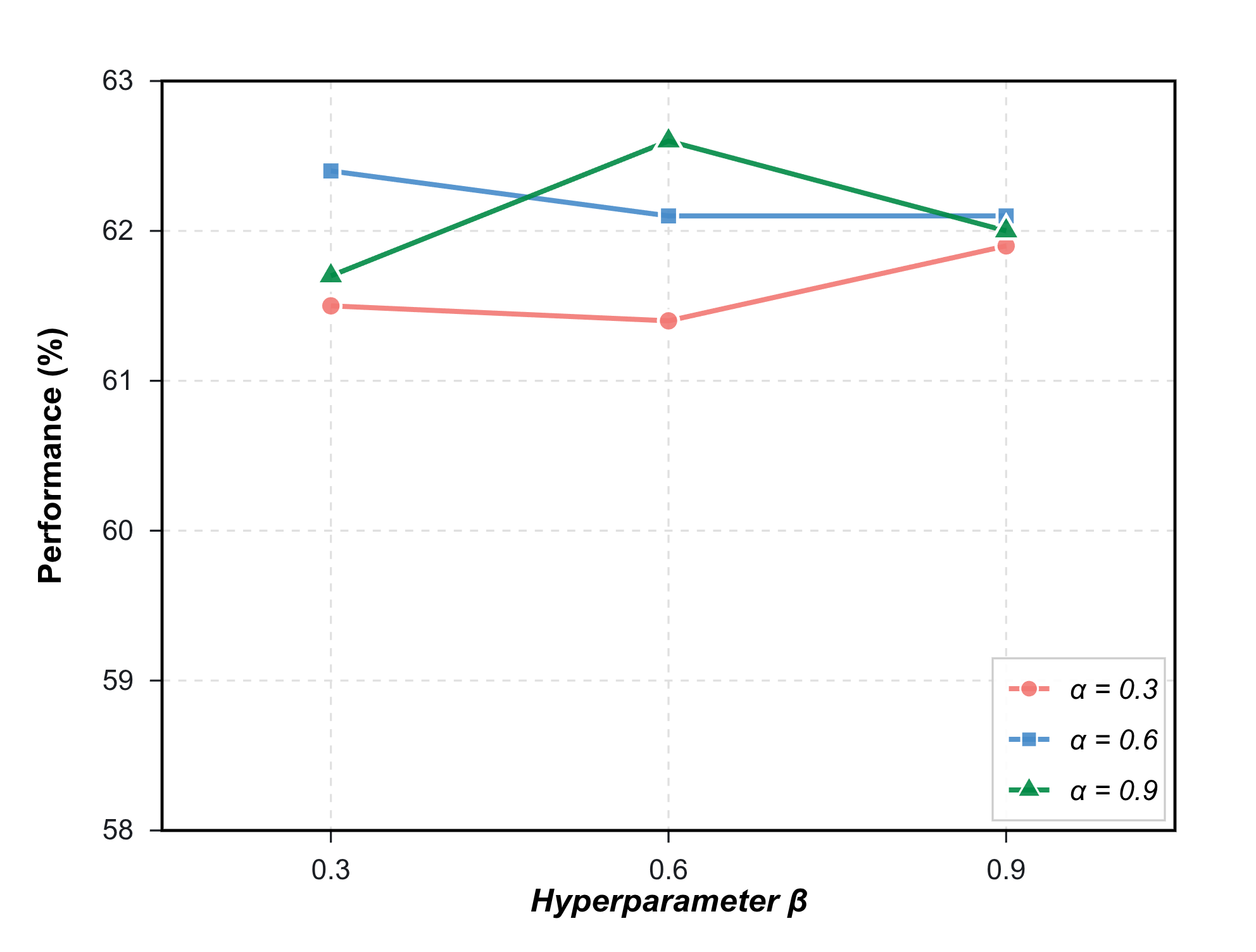}
  \caption{Sensitivity of the Dempster-Shafer hyperparameters $\alpha$ and $\beta$.}
  \label{fig:hyperparameter_sensitivity}
  \vspace{-1.0em}
\end{wrapfigure}

We further analyze the sensitivity of BayesRAG to the Dempster-Shafer trust parameters $\alpha$ and $\beta$. They control how strongly each modality contributes positive and negative evidence during fusion. Figure~\ref{fig:hyperparameter_sensitivity} reports the LLM score on MMLongBench-Doc with Qwen3-VL-32B as the generator. We evaluate all nine combinations of $\alpha,\beta\in\{0.3,0.6,0.9\}$.

Across these settings, scores remain within 61.4\%--62.6\%, a range of 1.2 percentage points. Moderate values already perform competitively: $\alpha=0.6,\beta=0.6$ yields 62.1\%, close to neighboring configurations. The highest grid value is 62.6\% at $\alpha=0.9,\beta=0.6$, while $\alpha=0.6,\beta=0.3$ gives 62.4\%. Performance therefore changes modestly across the evaluated combinations, rather than depending on a single narrow choice. We retain $\alpha=0.7,\beta=0.6$ for the main experiments.

\subsection{Efficiency and Computational Analysis}
\label{sec:efficiency}

Table~\ref{tab:time_cost} reports the recorded per-query stage averages for retrieval, reranking, fusion, and generation. The profile is collected on a server with NVIDIA Tesla A800 80GB GPUs and a 12-core CPU.

\begin{wraptable}{r}{0.4\textwidth}
  % \vspace{-1.0em}
  \centering
  \scriptsize
  \setlength{\tabcolsep}{4pt}
  \renewcommand{\arraystretch}{0.95}
  \caption{Average time cost (s) per query.}
  \begin{tabular}{@{}lc@{}}
    \toprule
    \textbf{Module} & \textbf{Time(s)} \\ \midrule
    Vector Retrieval (Text) & 0.43 \\
    Vector Retrieval (Image) & 5.50 \\
    Vector Retrieval (Screenshot) & 8.51 \\ \midrule
    Text Reranking (bge-reranker-v2-m3) & 1.63 \\
    \textbf{Bayes Fusion} & \textbf{1.91} \\ \midrule
    LLM Generation & 32.68 \\ \bottomrule
  \end{tabular}
  \label{tab:time_cost}
  % \vspace{-1.0em}
\end{wraptable}

\paragraph{Retrieval Latency vs. Recall.}
Image and screenshot retrieval together account for 14.01s per query, compared with 0.43s for text retrieval. These stages construct the visual candidate pools used by posterior reranking. BayesRAG retrieves 512 candidates per modality before selecting 20 items from each modality for generation. The broad initial pools retain candidates below the leading embedding matches, which can then be assessed using cross-modal support.

\paragraph{Efficiency of Bayesian Inference.}
The Bayes Fusion stage takes 1.91s per query, alongside 1.63s for text reranking. It scores the sparse tuple set constructed by the procedure in Section~\ref{sec:evidence_selection}. The recorded fusion cost is small relative to visual retrieval and the 32.68s required for answer generation. Appendix~\ref{app:tuple_construction} details how candidate construction restricts the tuple set before scoring.

\subsection{Conflict Frequency and Posterior Reranking}
\label{sec:retrieval_diagnostics}

\begin{wraptable}{r}{0.48\textwidth}
  % \vspace{-1em}
  \centering
  \caption{Conflict frequency and posterior reranking (\%).}
  \label{tab:conflict_diagnostics}
  \scriptsize
  \setlength{\tabcolsep}{3pt}
  \renewcommand{\arraystretch}{0.95}
  \begin{tabular*}{\linewidth}{@{\extracolsep{\fill}}lr@{}}
    \toprule
    \textbf{Statistic} & \textbf{Fraction} \\ \midrule
    Candidate tuples with $K>0.3$ & 63.5 \\
    Candidate tuples with $K>0.5$ & 26.2 \\
    Queries with changed Top-20 evidence sets & 85.7 \\
    \bottomrule
  \end{tabular*}
\end{wraptable}

Table~\ref{tab:conflict_diagnostics} reports conflict among modality-specific relevance signals and changes in the selected Top-20 evidence.
Among candidate tuples, 63.5\% have $K>0.3$, and 26.2\% have $K>0.5$, showing that opposing relevance signals occur frequently within the retrieved pool. Compared with embedding-similarity ranking, BayesRAG's posterior reranking changes at least one item in the final Top-20 evidence set for \textbf{85.7\%} of queries. The posterior score therefore changes the evidence supplied to the generator across most queries.

\section{Related Works}
\label{sec:related_works}

\noindent\textbf{Multimodal Vector Retrieval Paradigms.} RAG significantly enhances LLMs by integrating external knowledge bases~\citep{lewis2020retrieval}, with embedding models~\citep{multi2024chen} playing a pivotal role in this process. Recently, vision-language models such as CLIP and ColPali~\citep{faysse2024colpali} have extended RAG capabilities to retrieve images or document pages directly. Building on these advancements, RegionRAG~\citep{li2025regionragregionlevelretrievalaugmentedgeneration} introduces a region-level contrastive learning objective, thereby refining retrieval granularity from coarse-grained whole pages to specific semantic regions. Broadly, current approaches fall into two categories: (1) Visual-Centric Unification: Methods like VisRAG~\citep{yu2025visrag} and M3DocRAG~\citep{cho2024m3docrag} unify modalities by treating entire document pages as images for retrieval, effectively preserving layout but potentially sacrificing fine-grained textual details. (2) Dual-Stream Retrieval: Systems like VisDoM~\citep{suri2025visdom} store and retrieve text and images separately. They typically employ a late fusion strategy, where separate visual and textual models generate intermediate answers that are subsequently merged by a fusion model. However, these separate streams often lack interaction during the retrieval phase, leading to potential semantic misalignment.

\noindent\textbf{Knowledge-Structured Multimodal RAG.} To capture complex relationships within documents, several approaches integrate Knowledge Graphs. Methods such as HM-RAG, MMGraphRAG~\citep{wan2025mmgraphrag}, and mKG-RAG~\citep{yuan2025mkg} utilise Multimodal LLMs (MLLMs) to extract entities and relationships from both text and images. RAG-Anything~\citep{guo2025rag} employs a hybrid search mechanism that combines explicit KG links with vector semantic similarity. It utilizes a multi-signal scoring mechanism that weighs both the topological importance of the graph structure and the semantic similarity scores. While effective, these graph-based methods rely heavily on the quality of upstream information extraction.

\noindent\textbf{Evidence Verification and Reranking.} Ensuring the reliability of retrieved content is critical. For text modality, cross-encoder rerankers are standard practice. MIRAGE~\citep{wei2025mirage} ensures trustworthy evidence by cross-validation of inference chains in enhanced retrieval-augmented graphs. For images, frameworks like ViDoRAG~\citep{wang2025vidorag}, MDocAgent~\citep{han2025mdocagent}, HKRAG~\citep{tong2025hkragholisticknowledgeretrievalaugmented}, mRAG~\citep{hu2025mragelucidatingdesignspace}, and MADAM-RAG~\citep{wang2025retrieval} adopt an agent-based approach, utilizing agents such as the Inspector or Critical Agent to perform iterative verification loops to filter evidence. However, these agent-based methods are often computationally intensive due to repeated LLM calls, resulting in substantial token consumption and high latency. Furthermore, they lack a unified probabilistic framework to explicitly model the mutual corroboration between heterogeneous modalities—a gap we aim to fill with BayesRAG.

\section{Conclusion}
\label{sec:conclusion}

In this work, we introduce BayesRAG to advance multimodal RAG for visually rich documents by transforming retrieval into a dynamic evidence fusion process. By synergizing Bayesian inference with Dempster-Shafer theory, our framework offers a theoretically rigorous solution to the semantic inconsistencies prevalent in heterogeneous data. We demonstrate that explicitly modeling modal conflicts allows the system to filter out high-scoring but semantically misaligned noise. This capability effectively closes the gap between visual retrieval and semantic verification. Our extensive evaluations confirm that this probabilistic approach not only achieves state-of-the-art performance but also establishes a new paradigm for trustworthy multimodal RAG. We hope this shift towards evidence-based corroboration paves the way for future systems that demand high-fidelity reasoning.

\begin{ack}
  This work was supported by the Natural Science Foundation of China (62276242), Hefei Municipal Natural Science Foundation (HZR2431), CAAI-MindSpore Open Fund, developed on OpenI Community, the National Natural Science Foundation of China under Grant No. 62676401, the Beijing Nova Program under Grant 20250484899 and National Natural Science Foundation of China under Grant No. 62536009.
  This work was also supported by Graduate School of USTC under Grant No. BX-2025-1Y-014.
\end{ack}

\medskip

\bibliographystyle{unsrtnat}
\bibliography{main}

%%%%%%%%%%%%%%%%%%%%%%%%%%%%%%%%%%%%%%%%%%%%%%%%%%%%%%%%%%%%

\appendix

\clearpage

\section{Implementation of Bayesian inference}

\subsection{Likelihood Estimation via Dempster-Shafer Theory}
\label{app:likelihood_estimation}

In this appendix, we detail the calculation of the likelihood term $P(Q|E)$, which approximates the semantic support of the evidence tuple $E = (t,v,s)$ for the query $Q$. We utilize Dempster-Shafer Theory to robustly fuse heterogeneous retrieval scores and explicitly model uncertainty.

\subsubsection{Score Normalization and Sharpening}
Since retrieval scores from different embedding models (e.g., dense text embeddings vs. visual embeddings) often inhabit different numerical ranges, direct combination is infeasible. We first apply a normalization function to map raw scores into a unit interval $[0, 1]$.

For a raw score $s$ from modality $i$, given the empirically observed maximum ($s_{max}$) and minimum ($s_{min}$) scores for that modality, the normalized score $\tilde{s}_i$ is computed as:
\begin{equation}
  \tilde{s}_i = \frac{s - s_{min}}{s_{max} - s_{min}}
\end{equation}

\subsubsection{Mass Function Construction}
\label{app:mass_function}
We define the \textit{Frame of Discernment} as $\Theta = \{Y,N\}$, representing whether the evidence is ``Relevant'' ($Y$) or ``Irrelevant'' ($N$). For each normalized score $\tilde{s}_i$, we construct a Basic Probability Assignment, denoted as mass function $m_i(\cdot)$, based on hyperparameters $\alpha$ and $\beta$ (representing our trust in the retriever's positive and negative capabilities):

\begin{align}
  m_i(Y) &= \alpha \cdot \tilde{s}_i \\
  m_i(N) &= \beta \cdot (1 - \tilde{s}_i) \\
  m_i(\Theta) &= 1 - m_i(Y) - m_i(N)
\end{align}
Here, $m_i(\Theta)$ represents the \textit{uncertainty} (or ignorance) remaining after observing the score. We enforce $m_i(\Theta) \ge 0$ by clipping. Based on experimental experience, we fix $\alpha=0.7$ and $\beta=0.6$, respectively, for the main configuration.

\subsubsection{Recursive Evidence Combination}
To fuse evidence from multiple sources (Text, Image, Screenshot), we apply Dempster's Rule of Combination recursively. Let $m_{curr}$ be the accumulated mass and $m_{new}$ be the mass of the next incoming modality. The combined mass $m_{comb} = m_{curr} \oplus m_{new}$ is calculated as follows:

First, we compute the conflict coefficient $K$, which measures the degree of contradiction between the two evidence sources:
\begin{equation}
  \begin{split}
    K = m_{curr}(Y) \cdot  m_{new}(N) + m_{curr}(N) \cdot  m_{new}(Y)
  \end{split}
\end{equation}

If $K$ exceeds a threshold, we set the likelihood score to 0. Otherwise, we proceed with the update rules:
\begin{equation}
  \begin{split}
    m_{comb}(H) = \frac{\sum_{A \cap B = H} m_{curr}(A) \cdot m_{new}(B)}{1 - K}, \quad \varnothing\ne H\subseteq\Theta,\quad A,B\subseteq\Theta
  \end{split}
\end{equation}
Specifically for our binary frame, the update logic for the relevant hypothesis $Y$ is:
\begin{equation}
  \begin{split}
    m_{comb}(Y) = \frac{1}{1 - K} \Big( & m_{curr}(Y) \cdot m_{new}(Y) \\
      & + m_{curr}(Y) \cdot m_{new}(\Theta) \\
    & + m_{curr}(\Theta) \cdot m_{new}(Y) \Big)
  \end{split}
\end{equation}

\subsubsection{Final Score}
After fusing all evidence sources, we obtain the final mass distribution $m_{final}$. To convert this belief function into a decision score suitable for ranking (i.e., $P(Q|E)$), we apply the \textit{Pignistic Probability Transformation}, which distributes the uncertainty mass $\Theta$ equally among the singleton hypotheses:
\begin{equation}
  \begin{split}
    P(Q|E) \approx \text{BetP}_{joint}(Y) = m_{final}(Y) + \frac{m_{final}(\Theta)}{2}
  \end{split}
\end{equation}
This final score reflects a decision-level ranking quantity, accounting for both the strength of individual matches and the consistency across modalities.

We provide the pseudocode for the likelihood estimation process $P(Q|E)$ in Algorithm~\ref{alg:ds_likelihood}.

\begin{algorithm}[htp]
  \caption{Likelihood Estimation via Dempster-Shafer Fusion}
  \label{alg:ds_likelihood}
  \begin{algorithmic}[1]
    \REQUIRE Raw retrieval scores $\mathcal{S} = \{s_{txt}, s_{img}, s_{screenshot}\}$
    \REQUIRE Hyperparameters $\alpha, \beta \in [0, 1]$ (Trust factors)
    \REQUIRE Normalization function $\Phi(\cdot)$
    \ENSURE Likelihood score $P(Q|E)$

    \STATE \textbf{Initialize:}
    \STATE $m_{curr} \leftarrow \{Y: 0, N: 0, \Theta: 1\}$ \COMMENT{Initial state: total ignorance}

    \STATE \textbf{Preprocessing:}
    \STATE $\mathcal{S}' \leftarrow [\Phi(s) \text{ for } s \text{ in } \mathcal{S}]$ \COMMENT{Normalize scores to $[0,1]$}

    \FOR{each score $s \in \mathcal{S}'$}
    \STATE \textbf{Construct Mass Function (BPA):}
    \STATE $m_{new}(Y) \leftarrow \alpha \cdot s$
    \STATE $m_{new}(N) \leftarrow \beta \cdot (1 - s)$
    \STATE $m_{new}(\Theta) \leftarrow \max(0, 1 - m_{new}(Y) - m_{new}(N))$

    \STATE \textbf{Compute Conflict Coefficient $K$:}
    \STATE $K \leftarrow m_{curr}(Y) \cdot m_{new}(N) + m_{curr}(N) \cdot m_{new}(Y)$

    \IF{$K \ge threshold$}
    \RETURN $0.0$ \COMMENT{Extreme conflict detected}
    \ENDIF

    \STATE \textbf{Apply Dempster's Rule:}
    \STATE $D \leftarrow 1 - K$
    \STATE $m_{next}(Y) \leftarrow (m_{curr}(Y) \cdot m_{new}(Y) + m_{curr}(Y) \cdot m_{new}(\Theta) + m_{curr}(\Theta) \cdot m_{new}(Y)) / D$
    \STATE $m_{next}(N) \leftarrow (m_{curr}(N) \cdot m_{new}(N) + m_{curr}(N) \cdot m_{new}(\Theta) + m_{curr}(\Theta) \cdot m_{new}(N)) / D$
    \STATE $m_{next}(\Theta) \leftarrow (m_{curr}(\Theta) \cdot m_{new}(\Theta)) / D$

    \STATE $m_{curr} \leftarrow m_{next}$
    \ENDFOR

    \STATE \textbf{Final Score:}
    \STATE $Score \leftarrow m_{curr}(Y) + m_{curr}(\Theta) / 2$

    \RETURN $Score$
  \end{algorithmic}
\end{algorithm}

\subsection{Evidence-Tuple Construction and Context Selection}
\label{app:tuple_construction}

Text chunks, locally cropped visual elements, and full-page screenshots are retrieved from separate streams, with an initial Top-512 pool for each modality. Each scored evidence tuple contains one candidate from each stream. We construct these tuples after retrieval.

Combining every text, image, and screenshot candidate from the three Top-512 pools would yield $512^3\approx134$ million tuples. BayesRAG uses a sparse candidate-construction strategy before calculating posterior scores. For each modality, we retain the top-$h$ candidates by embedding score as a core pool and fully enumerate their combinations, producing $h^3$ core tuples. Candidates outside the core are admitted through compatible cross-modal combinations. In the collection-level setting, a same-document constraint excludes combinations whose components originate from different documents. For example, a text chunk from one document is never paired with an image from another. In the single-document setting, this condition is already satisfied by the initial \texttt{doc\_id} filter.

We then apply knowledge-graph filtering to the surviving tuples. Their source regions are mapped to graph nodes, and tuples with weak connectivity are pruned before scoring. The core pool retains the leading embedding matches, while the compatibility and graph checks restrict the additional combinations considered from the tail. Together, these steps provide a practical way to use a broad initial pool while limiting the candidate set evaluated by Bayesian fusion.

Tuples are sorted by the product of their likelihood and consistency scores, following Section~\ref{sec:evidence_selection}. The top 20 items from each modality are then used as the final input to the generator. Thus, each stream begins with 512 candidates and contributes 20 selected items to the context.

\subsection{Implementation of Priors}
\label{app:priors_implementation}

We detail the exact algorithms used to compute the structural prior $P(E)$ for both document layout and knowledge graph scenarios. The prior $P(E)$ measures the intrinsic coherence of an evidence tuple $E = (e_{txt}, e_{img}, e_{screenshot})$ derived from geometric or semantic constraints.

\subsubsection{Layout-Aware Prior Calculation}
\label{app:layout_prior}

For document-native formats (e.g., PDFs), we utilize bounding box (BBox) coordinates to determine spatial proximity. The calculation of $P(E)_{\text{layout}}$ is implemented as a strict geometric consistency check.

Given a text node $t$, an image node $v$, and a full-page screenshot node $s$, we first verify the consistency of the metadata. The tuple is considered valid only if all three components originate from the same source document $D$. Let $pg(\cdot)$ denote the page index and $d(\cdot, \cdot)$ denote the Euclidean distance between the centers of two bounding boxes.

For elements on the same page, $d(t,v)=\|c_t-c_v\|_2$, where $c_t$ and $c_v$ are bounding-box centers in the MinerU page coordinate system. We normalize by $L_{diag}=\sqrt{W^2+H^2}$ for page width $W$ and height $H$. For adjacent pages, we concatenate the full pages into a shared canvas and transform the bounding-box centers into that coordinate system before measuring their distance. This preserves proximity across a page boundary.

The prior is computed as a binary indicator function with a relaxation for adjacent pages:
\begin{equation}
  P(E)_{\text{bbox}} =
  \begin{cases}
    1.0 & \text{if } \mathcal{C}_{\text{dist}} \land \mathcal{C}_{\text{page}} \\
    \epsilon & \text{otherwise}
  \end{cases}
\end{equation}
where $\epsilon$ is a small constant (e.g., $\epsilon=0.1$). The conditions are defined as:

\begin{enumerate}
  \item \textbf{Spatial Condition} ($\mathcal{C}_{\text{dist}}$): The spatial distance must be within a fraction of the page diagonal length $L_{diag}$:
    \begin{equation}
      d(t, v) < \tau \cdot L_{diag}
    \end{equation}
    where $\tau$ is a distance threshold parameter (e.g., $\tau=2$).

  \item \textbf{Pagination Condition} ($\mathcal{C}_{\text{page}}$): The elements must appear on the same or immediately adjacent pages relative to the screenshot reference:
    \begin{equation}
      \max(|pg(t) - pg(s)|, |pg(v) - pg(s)|) < \tau_{page}
    \end{equation}
\end{enumerate}
This implementation ensures that a high prior probability is assigned solely to figure-caption pairs or spatially clustered information.

\subsubsection{Knowledge Graph Topology Prior}
\label{app:kg_prior}

For the knowledge graph, we construct a weighted connectivity map and compute the coherence of the retrieved triplet. This process consists of three steps: Relation Aggregation, Edge Normalization, and Triplet Coherence Scoring.

\textbf{Step 1: Relation Aggregation}.
The nodes correspond to retrieved text chunks, visual regions, and page screenshots. Entities extracted from document content are mapped to their source regions, and graph relations induce connections between these nodes. For any pair of chunks $(u, v)$, we aggregate the weights of all relations $r$ connecting them in the Knowledge Graph:
\begin{equation}
  S_{uv} = \sum_{r \in \mathcal{R}_{u,v}} w_r
\end{equation}
where $w_r$ is the weight of a specific relation (defaulting to 1.0 or weighted by relation type).

\textbf{Step 2: Probabilistic Edge Normalization}.
To convert the unbounded raw score $S_{uv}$ into a probability $P_{edge}(u, v) \in [0, 1)$, we apply an exponential saturation function. This models the intuition that diminishing returns apply as evidence accumulates:
\begin{equation}
  P_{edge}(u, v) = 1 - \exp(-\kappa \cdot S_{uv})
\end{equation}
where $\kappa$ is a scaling factor (set to 0.1 in our experiments). This defines the pairwise connectivity map.

\textbf{Step 3: Triplet Coherence Scoring}.
For a candidate tuple $E = (e_{id_1}, e_{id_2}, e_{id_3})$, we retrieve the three pairwise edge probabilities: $p_{12}, p_{23}, p_{13}$. We calculate their average as the final prior score

\begin{equation}
  \begin{split}
    P(E)_{\text{graph}} = \frac{p_{12}+p_{23}+p_{13}}{3}
  \end{split}
\end{equation}

\section{Reproducibility}
\label{app:reproducibility}

In this section, we provide detailed configurations for the baseline models to ensure reproducibility. For most parameters, we use the default settings in the original repository. Across all RAG systems, we employ identical embedding and reranker models.

\subsection{VisRAG}
We follow to the reproducible baseline methodology provided by UltraRAG\footnote{\url{https://github.com/OpenBMB/UltraRAG}}. We adapt the data structures of MMLongBench-Doc and DocBench to align with the benchmark schema required by UltraRAG. Question answering is conducted using the official configuration template\footnote{\url{https://github.com/OpenBMB/UltraRAG/blob/main/examples/visrag.yaml}}. Finally, answer extraction and scoring are performed using the evaluation scripts provided by the respective benchmarks.

\subsection{ViDoRAG}
We follow the official implementation of ViDoRAG\footnote{\url{https://github.com/Alibaba-NLP/ViDoRAG}} with specific modifications to the embedding module. Specifically, we utilize \texttt{MinerU} for document parsing. For vector database construction, we employ \texttt{Qwen3-Embedding-4B} for text segments and \texttt{colnomic-embed-multimodal-3b} for page-level PDF screenshots. We apply the proposed Gaussian Mixture Model (GMM)-based hybrid strategy to effectively handle multi-modal retrieval. \texttt{Qwen3-VL-32B} serves as the backbone model for the Seeker, Inspector, and Answer agents.

\subsection{RAGFlow}
We deploy RAGFlow using the official Docker image\footnote{\url{https://hub.docker.com/r/infiniflow/ragflow}}. We employ \texttt{Qwen3-Embedding-4B} and \texttt{bge-reranker-v2-m3} (host via VLLM) for text embedding and reranking, respectively. For multimodal content, \texttt{Qwen3-VL-32B} is utilized to generate descriptions. All other parsing parameters retain their default settings.

During the question-answering evaluation, we use \texttt{Qwen3-VL-32B} and enable the following features: \textit{Keyword Analysis}, \textit{TOC Enhance}, \textit{Use Knowledge Graph}, and \textit{Reasoning}. The specific chat configuration parameters are detailed in Listing~\ref{lst:ragflow_config}.

\clearpage

\begin{lstlisting}[
    language=json,
    caption={Configuration parameters for RAGFlow.},
    label={lst:ragflow_config},
    breakindent=20pt,
    postbreak=\mbox{\textcolor{gray}{$\hookrightarrow$}\space} % 换行指示符
]
{
  "language": "English",
  "llm": {
    "max_tokens": 4096,
    "model_name": "Qwen3-VL-32B___OpenAI-API@OpenAI-API-Compatible",
    "temperature": 0.1
  },
  "meta_data_filter": {
    "method": "automatic"
  },
  "prompt": {
    "empty_response": "",
    "keyword": true,
    "keywords_similarity_weight": 0.7,
    "opener": "Hi! I'm your assistant. What can I do for you?",
    "prompt": "You are an intelligent assistant. Please summarize the content of the knowledge base to answer the question. Please list the data in the knowledge base and answer in detail. When all knowledge base content is irrelevant to the question, your answer must include the sentence \'The answer you are looking for is not found in the knowledge base!\' Answers need to consider chat history.\n      Here is the knowledge base:\n      {knowledge}\n      The above is the knowledge base.",
    "reasoning": true,
    "refine_multiturn": false,
    "rerank_model": "bge-reranker-v2-m3___OpenAI-API@OpenAI-API-Compatible",
    "show_quote": true,
    "similarity_threshold": 0.2,
    "toc_enhance": true,
    "top_n": 8,
    "tts": false,
    "use_kg": true,
    "variables": [
      {
        "key": "knowledge",
        "optional": true
      }
    ]
  },
  "prompt_type": "simple",
  "top_k": 1024
}
\end{lstlisting}

\subsection{RAGAnything}
Consistent with our other setups, we employ \texttt{MinerU} as the document parser. We utilize \texttt{Qwen3-VL-32B} to generate document descriptions and construct the knowledge graph. The vector retrieval component relies on \texttt{Qwen3-Embedding-4B} and \texttt{bge-reranker-v2-m3}. Finally, \texttt{Qwen3-VL-32B} is employed as the downstream question-answering model.

\subsection{Compute Requirements}
Following the setup in Section~\ref{sec:efficiency}, our measurements are obtained on a server equipped with NVIDIA A800 80GB GPUs and a 12-core CPU. The sum of the recorded stage averages is approximately 50.66s per query, including text retrieval (0.43s), image retrieval (5.50s), screenshot retrieval (8.51s), text reranking (1.63s), Bayes fusion (1.91s), and final generation (32.68s).

Based on the benchmark sizes, a single serial evaluation pass requires approximately 15 hours on MMLongBench-Doc, 16 hours on DocBench, and about 31 hours for both benchmarks combined.

\section{Prompts in BayesRAG}
\label{app:prompts}

The specific prompts employed in BayesRAG are detailed below:

\begin{figure}[h]
    \begin{tcolorbox}[
        colback=white,
        colframe=black,
        title=Prompt
    ]
    You are an intelligent assistant. Please summarize the content of the knowledge base (both in text, images, and screenshots of pdf pages) to answer questions, think step by step. \\
    
    First, analyze the keywords in the question, and then check whether all texts and images of the given knowledge base contain content related to the keywords in the question. Finally summarize and answer questions. \\
    
    When all knowledge base content is not related to the question, your answer should include the sentence ``The answer you want is not found in the knowledge base!'' \\
    
    Here is the question: \\

    Question: \{question\} \\

    The above is the question. \\
    
    Here is the knowledge base: \\

    \{knowledge\} \\

    The above is the knowledge base.
    \end{tcolorbox}
    \caption{Prompts used by the model to predict answers.}
    \label{fig:prompt_answer}
\end{figure}

\begin{figure}[h]
    \begin{tcolorbox}[
        colback=white,
        colframe=black,
        title=Prompt
    ]
    You are required to determine if a predicted answer is correct or can reasonably answer the question compared to the ground truth. The question will be placed within <question></question> tags, answer type will be placed within <type></type> tags, predicted answer will be placed within <predict></predict> tags, and the ground truth answer will be placed within <gt></gt> tags. \\

    You must output the final score as a floating-point number (0.0 to 1.0) enclosed within `<answer>' tags. \\
    
    Output Format: Output \textbf{only} the final numerical score inside the tags. Do not provide reasoning. \\
    
    Example: \\
    <answer> 1.0 </answer> \\
    
    Question: <question> \{question\} </question> \\
    
    Predict Answer: <predict> \{predict answer\} </predict> \\
    
    Ground Truth: <gt> \{reference answer\} </gt>
    \end{tcolorbox}
    \caption{Prompts used by evaluating the answer in MMLongBench-Doc.}
    \label{fig:prompt_gpt_score}
\end{figure}
\clearpage
\section{Case Study}
\label{sec:case_study}

\begin{figure*}[tp]
  % \vspace{-5mm}
  \centering
  \includegraphics[width=\textwidth]{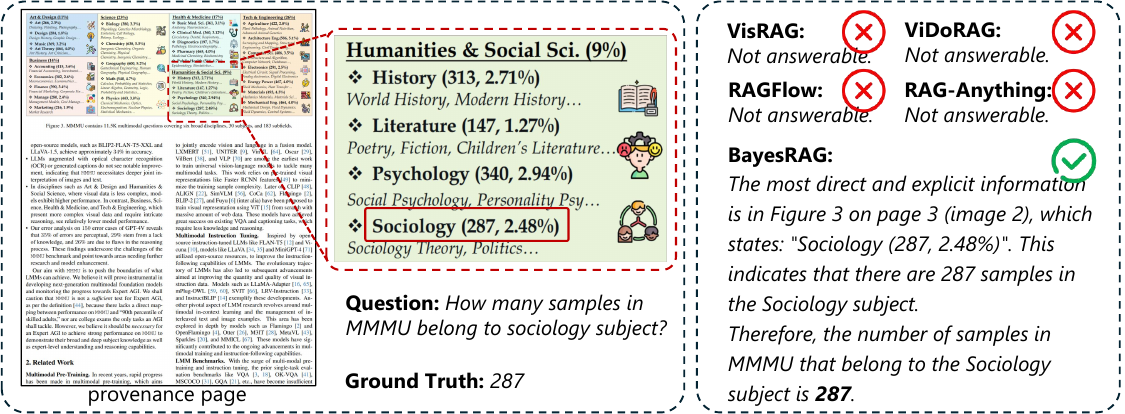}
  \caption[Case 1]
  {
    \textmd{A qualitative comparison illustrating the efficacy of BayesRAG in broadening the scope of visual retrieval and filtering inconsistent evidence. Our statistical analysis reveals that approximately 34\% of failure cases in baseline methods are attributed to this retrieval deficiency.}
  }
  \label{fig:case1}
  \vspace{-3mm}
\end{figure*}

\begin{figure*}[tp]
  % \vspace{-5mm}
  \centering
  \includegraphics[width=\textwidth]{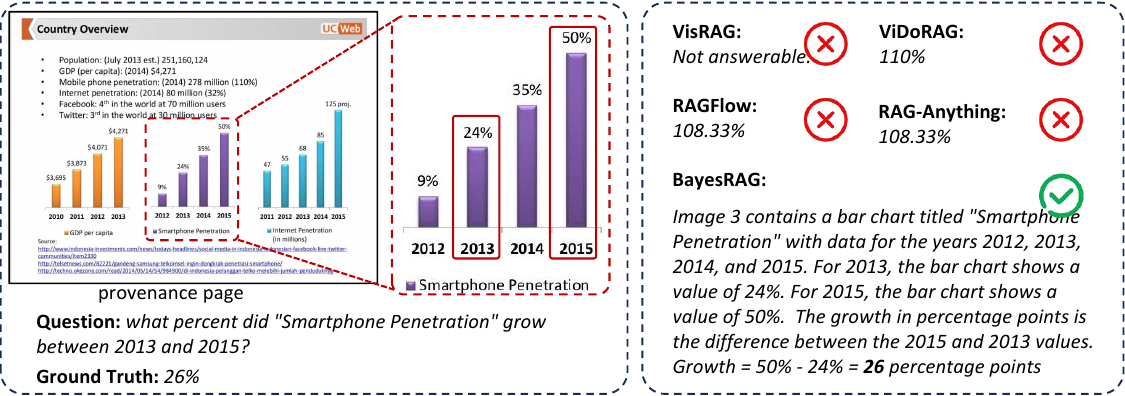}
  \caption[Case 2]
  {
    \textmd{A sample is for semantic consistency between text and charts to find the correct evidence.}
  }
  \label{fig:case2}
  \vspace{-5mm}
\end{figure*}

\begin{figure*}[tp]
  % \vspace{-5mm}
  \centering
  \includegraphics[width=\textwidth]{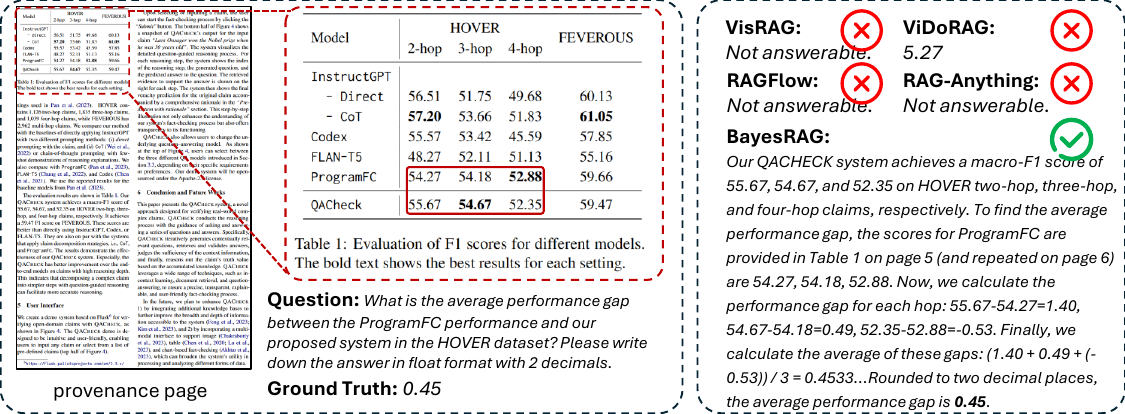}
  \caption[Case 3]
  {
    \textmd{Complex tabular reasoning requiring multi-hop numerical extraction.}
  }
  \label{fig:case3}
\end{figure*}

\begin{figure*}[tp]
  % \vspace{-5mm}
  \centering
  \includegraphics[width=\textwidth]{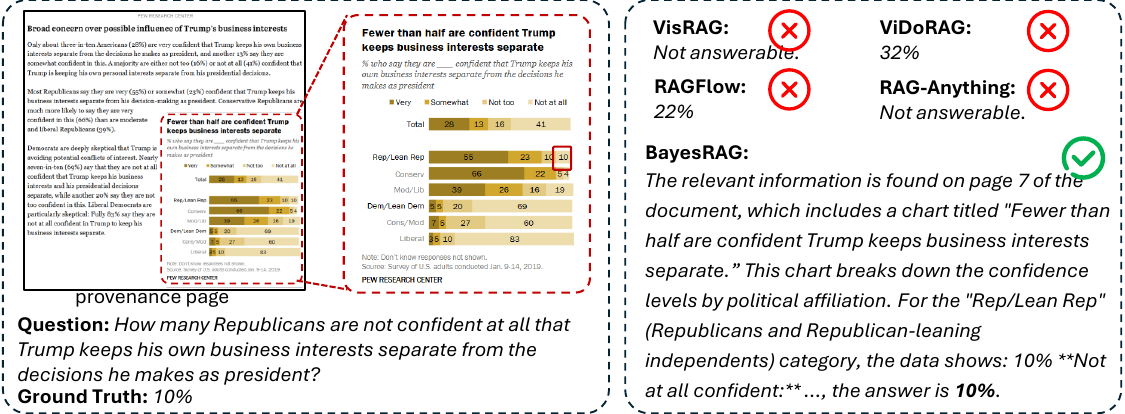}
  \caption[Case 4]
  {
    \textmd{Fine-grained visual-textual alignment in stacked bar charts.}
  }
  \label{fig:case4}
\end{figure*}

\begin{figure*}[tp]
  % \vspace{-5mm}
  \centering
  \includegraphics[width=\textwidth]{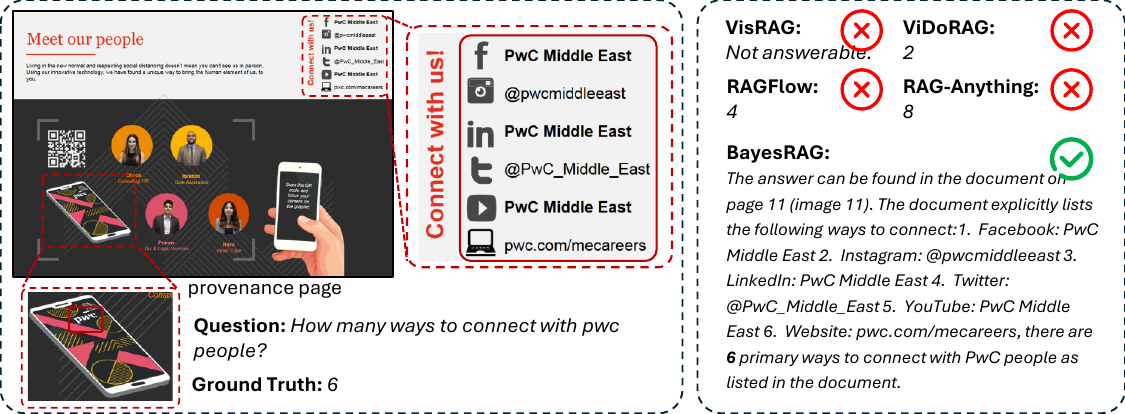}
  \caption[Case 5]
  {
    \textmd{Information enumeration from unstructured infographics.}
  }
  \label{fig:case5}
\end{figure*}

To explicitly demonstrate how BayesRAG handles complex retrieval challenges, we present two representative cases in Figure~\ref{fig:case1} and Figure~\ref{fig:case2}, followed by three additional qualitative examples in Figure~\ref{fig:case3}, Figure~\ref{fig:case4}, and Figure~\ref{fig:case5}.

\noindent\textbf{Case 1:} Figure~\ref{fig:case1} depicts target text (``Sociology'') buried in a complex layout. Baselines fail to capture this low-salience signal, returning ``Not answerable''. BayesRAG addresses this by expanding the retrieval candidate pool (Top-$k$) and leveraging the semantic consistency between the query and the image text to filter the resulting noise effectively. This capability is critical, as our analysis reveals that about 34\% of baseline errors stem from similar failures to retrieve ground-truth evidence, directly resulting in incorrect ``Not answerable'' predictions.

\noindent\textbf{Case 2:} Figure~\ref{fig:case2} presents a challenge where baselines are misled by the visual similarity of a ``Mobile phone'' distractor chart located on a different slide. BayesRAG successfully retrieves the correct evidence through its probabilistic scoring mechanism. Specifically, the explicit match between the query ``Smartphone'' and the textual component of the target evidence tuple yields a higher \textbf{likelihood} score. Simultaneously, the semantic consistency between the tuple's text and image results in a robust \textbf{prior} score. These factors synergize during the Bayesian update to rank the correct ``Smartphone'' tuple significantly higher than the visually similar distractor, enabling the accurate calculation.

In summary, BayesRAG not only expands recall for obscure details in images but also robustly filters confusing distractors through rigorous probabilistic verification.

Beyond these two representative examples from the main analysis, we provide three additional qualitative cases highlighting a critical advantage of our framework: the ability to solve complex, multi-hop reasoning tasks where baseline methods fail to retrieve the correct evidence entirely.

A common failure mode in existing multimodal RAG systems is the inability to locate evidence that is intricately embedded within structured documents, such as complex tables, stacked charts, or dense infographics. Single-stream retrievers (relying solely on text chunks or naive visual embeddings) often miss these cues due to a lack of fine-grained semantic alignment between the query and the visual structure.

BayesRAG addresses this through its parallel retrieval architecture, simultaneously querying text chunks, cropped image regions, and full-page PDF screenshots. Crucially, the subsequent Bayesian fusion mechanism calculates a posterior evidence score combining query relevance with consistency among these heterogeneous evidence sources. This allows BayesRAG to identify and utilize evidence that requires establishing strong cross-modal links (e.g., mapping legend colors to textual categories, or aligning table headers with specific numerical cells), tasks where baselines frequently hallucinate or fail to answer.

We present three such challenging scenarios below: complex tabular reasoning (Figure~\ref{fig:case3}), fine-grained chart interpretation (Figure~\ref{fig:case4}), and infographic enumeration (Figure~\ref{fig:case5}).

\section{Reliability of Automated Evaluation}
\label{app:reliability}

We assess Qwen3-32B as an automated judge on a stratified sample of 200 MMLongBench-Doc instances. The judge and expert human annotators agree on \textbf{98\%} of the sampled judgments. This agreement supports its use for answer evaluation in this benchmark. Many questions require specific numerical values or entities, making correspondence with the reference answer central to the scoring task.

The LLM judge also recognizes semantic equivalence when valid answers differ from the reference in wording or formatting. This complements the official rule-based accuracy by evaluating answer content beyond literal string matches.

\section{Collection-Level Evaluation}
\label{app:generator_impact}

We evaluate retrieval over the full benchmark collection with Qwen3-VL-32B and GPT-4o-mini as answering models. Each query searches the collection without a target-document filter, requiring the system to locate both the source document and supporting evidence.

\begin{table*}[ht]
  \caption{Collection-level LLM score (\%) on MMLongBench-Doc with Qwen3-VL-32B and GPT-4o-mini as answering models. Bold and underline indicate the best and second-best results within each answering-model block.}
  \label{tab:mmlongbench_results_diff_generater}
  \resizebox{\textwidth}{!}{%
    \begin{tabular}{@{}lcccccccc@{}}
      \toprule
      \textbf{Method} & \textbf{Reports} & \textbf{Tutorials} & \textbf{Academic} & \textbf{Guidebooks} & \textbf{Brochures} & \textbf{Industry} & \textbf{Financial} & \textbf{Overall} \\ \midrule
      \multicolumn{9}{l}{\textit{Answering model: Qwen3-VL-32B}} \\
      VisRAG & 45.4 & 46.0 & 35.3 & 49.4 & {\ul 43.6} & {\ul 44.4} & 40.2 & 43.4 \\
      ViDoRAG & 51.9 & 48.2 & 36.8 & 41.7 & 39.6 & 40.7 & 47.0 & 44.6 \\
      RAGFlow & 47.8 & {\ul 51.8} & 37.7 & 41.7 & 41.6 & 33.3 & {\ul 51.3} & 44.3 \\
      RAGAnything & {\ul 53.2} & 47.5 & {\ul 40.2} & {\ul 51.9} & 39.6 & 37.0 & 50.4 & {\ul 47.1} \\
      \rowcolor{green!12} BayesRAG (ours) & \textbf{61.8} & \textbf{57.6} & \textbf{58.3} & \textbf{59.0} & \textbf{54.5} & \textbf{54.3} & \textbf{65.0} & \textbf{59.3} \\
      \midrule
      \multicolumn{9}{l}{\textit{Answering model: GPT-4o-mini}} \\
      GPT-4o-mini (direct input) & 36.7 & 38.2 & 24.0 & 32.4 & 34.6 & \textbf{50.6} & 41.0 & 35.2 \\
      VisRAG & 30.0 & 20.8 & 24.0 & 23.7 & 23.7 & 24.6 & 11.1 & 23.8 \\
      ViDoRAG & 36.1 & 39.5 & 21.0 & 32.0 & 24.7 & 25.9 & 36.7 & 31.4 \\
      RAGFlow & 38.9 & 35.2 & 29.9 & 35.2 & 36.6 & 30.8 & 34.1 & 34.9 \\
      RAGAnything & {\ul 45.3} & {\ul 46.0} & {\ul 34.6} & {\ul 43.5} & \textbf{39.6} & 33.3 & {\ul 42.7} & {\ul 41.5} \\
      \rowcolor{green!12} BayesRAG (ours) & \textbf{46.7} & \textbf{53.6} & \textbf{34.8} & \textbf{46.4} & {\ul 39.4} & {\ul 38.6} & \textbf{48.6} & \textbf{44.1} \\
      \bottomrule
    \end{tabular}%
  }
\end{table*}

\begin{table*}[ht]
  \centering
  \caption{Collection-level accuracy (\%) on DocBench with Qwen3-VL-32B and GPT-4o-mini as answering models. Bold and underline indicate the best and second-best results within each answering-model block.}
  \label{tab:docbench_results_diff_generater}
    \begin{tabular}{@{}lcccccc@{}}
      \toprule
      \textbf{Method} & \textbf{Aca.} & \textbf{Fin.} & \textbf{Gov.} & \textbf{Law} & \textbf{News} & \textbf{Overall} \\ \midrule
      \multicolumn{7}{l}{\textit{Answering model: Qwen3-VL-32B}} \\
      VisRAG & {\ul 62.7} & {\ul 54.5} & 48.6 & 52.9 & 40.7 & {\ul 53.5} \\
      ViDoRAG & 47.9 & \textbf{56.3} & 35.8 & {\ul 57.6} & 50.0 & 50.5 \\
      RAGFlow & 50.2 & 46.9 & 31.8 & 48.7 & 36.0 & 44.4 \\
      RAGAnything & 57.4 & 44.8 & {\ul 49.3} & 50.8 & {\ul 61.0} & 52.5 \\
      \rowcolor{green!12} BayesRAG (ours) & \textbf{65.7} & 53.8 & \textbf{61.5} & \textbf{59.7} & \textbf{70.3} & \textbf{61.7} \\
      \midrule
      \multicolumn{7}{l}{\textit{Answering model: GPT-4o-mini}} \\
      GPT-4o-mini (direct input) & 42.5 & 36.8 & {\ul 53.1} & {\ul 51.4} & 51.1 & 44.1 \\
      VisRAG & 21.9 & 28.2 & 20.3 & 29.6 & 11.6 & 25.3 \\
      ViDoRAG & 29.0 & \textbf{69.1} & 35.8 & 39.7 & 37.2 & 43.5 \\
      RAGFlow & 45.2 & 34.0 & 41.8 & 48.1 & 23.8 & 39.0 \\
      RAGAnything & {\ul 51.4} & {\ul 42.3} & 43.2 & 50.2 & {\ul 57.5} & {\ul 48.7} \\
      \rowcolor{green!12} BayesRAG (ours) & \textbf{52.4} & 37.5 & \textbf{56.0} & \textbf{52.3} & \textbf{66.8} & \textbf{51.2} \\
      \bottomrule
    \end{tabular}
\end{table*}

\paragraph{Results with Qwen3-VL-32B.}
Tables~\ref{tab:mmlongbench_results_diff_generater} and~\ref{tab:docbench_results_diff_generater} report the collection-level results on MMLongBench-Doc and DocBench. With Qwen3-VL-32B as the answering model, BayesRAG achieves 59.3\% LLM score on MMLongBench-Doc, compared with 47.1\% for RAGAnything. On DocBench, BayesRAG reaches 61.7\% accuracy, compared with 52.5\% for RAGAnything. These correspond to gains of 12.2 and 9.2 percentage points, respectively, when retrieving evidence across multiple candidate documents.

\paragraph{Results with GPT-4o-mini.}
The GPT-4o-mini blocks include the direct-input baseline, VisRAG, ViDoRAG, RAGFlow, RAGAnything, and BayesRAG. With GPT-4o-mini as the answering model, BayesRAG and RAGAnything obtain 44.1\% and 41.5\% LLM score on MMLongBench-Doc, respectively. Their corresponding DocBench accuracies are 51.2\% and 48.7\%, giving BayesRAG gains of 2.6 and 2.5 percentage points across the two benchmarks.

For the direct-input GPT-4o-mini baseline, documents shorter than 50 pages use all page screenshots, while longer documents use 50 randomly sampled screenshots.

\paragraph{Results by Question Type.}
Table~\ref{tab:historical_docbench_types} reports DocBench accuracy for textual, metadata, multimodal, and unanswerable questions with each answering model. With Qwen3-VL-32B, BayesRAG achieves higher accuracy than RAGAnything in all four categories, including 62.6\% on multimodal questions and 39.5\% on unanswerable questions.

\begin{table}[ht]
  \centering
  \caption{Collection-level accuracy (\%) on DocBench by question type and answering model. Text, Meta, MM, and Unans. denote textual, metadata, multimodal, and unanswerable questions.}
  \label{tab:historical_docbench_types}
  \begin{tabular}{@{}lcccc@{}}
    \toprule
    Method & Text & Meta & MM & Unans. \\ \midrule
    \multicolumn{5}{l}{\textit{Answering model: Qwen3-VL-32B}} \\
    RAGAnything & 82.0 & 11.2 & 60.7 & 19.3 \\
    BayesRAG (ours) & 87.3 & 30.2 & 62.6 & 39.5 \\
    \midrule
    \multicolumn{5}{l}{\textit{Answering model: GPT-4o-mini}} \\
    RAGAnything & 77.6 & 10.0 & 57.4 & 11.2 \\
    BayesRAG (ours) & 84.2 & 19.7 & 43.8 & 25.8 \\
    \bottomrule
  \end{tabular}
\end{table}

\paragraph{Document Identification.}
A manual inspection of 100 failed MMLongBench-Doc cases found that 51\% were associated with uncertainty about the intended document during collection-level retrieval. Questions such as ``What year is the report for?'' can refer to several documents when the source identifier is withheld. Retrieving the correct page then requires resolving both which document is intended and which content supports the answer.

\section{Licenses for Existing Assets}
\label{app:licenses}

Table~\ref{tab:existing_asset_licenses} summarizes the major third-party datasets, codebases, models, and hosted services used in this work. We properly credit the original creators and use each asset in a manner consistent with its official license or terms of use. When the official release does not expose a clear license, we conservatively treat the asset as \emph{license-unclear} and do not redistribute it.
\clearpage

\begin{table*}[h]
  \centering
  \scriptsize
  \setlength{\tabcolsep}{4pt}
  \renewcommand{\arraystretch}{1.08}
  \caption{Third-party assets used in this paper and their licenses/terms of use.}
  \label{tab:existing_asset_licenses}
  \resizebox{\textwidth}{!}{%
    \begin{tabular}{@{}p{1.2cm}p{2.2cm}p{3.1cm}p{2.2cm}p{6.2cm}@{}}
      \toprule
      \textbf{Type} & \textbf{Asset} & \textbf{Role} & \textbf{Credit} & \textbf{License / Terms of Use} \\
      \midrule
      Data & DocBench & Evaluation benchmark & Zou et al.~\cite{zou2025docbench} & The official repository provides the benchmark description and a download link for the data, but does not expose an explicit dataset license in the release page. We therefore treat it as \emph{license-unclear}, use it only for academic evaluation, credit the original authors, and do not redistribute the benchmark files. \\
      Data & MMLongBench-Doc & Evaluation benchmark & Ma et al.~\cite{ma2024mmlongbench} & Apache-2.0. \\
      \midrule
      Code & RAGFlow & Baseline implementation & Infiniflow~\cite{infiniflow2024ragflow} & Apache-2.0. \\
      Code & VisRAG & Baseline implementation & Yu et al.~\cite{yu2025visrag} & Apache-2.0. \\
      Code & ViDoRAG & Baseline implementation & Wang et al.~\cite{wang2025vidorag} & The official repository does not expose an explicit license in the release page. We therefore treat it as \emph{license-unclear}, credit the authors, and avoid redistributing derived releases based on their code without further permission verification. \\
      Code & RAGAnything & Baseline implementation & Guo et al.~\cite{guo2025rag} & MIT. \\
      Code & MinerU & Document parsing & Niu et al.~\cite{niu2025mineru25decoupledvisionlanguagemodel} & Apache-2.0 with additional MinerU terms: commercial use is permitted without a separate license below the stated MAU/revenue thresholds, while large-scale deployments above those thresholds require a separate commercial license; online services must clearly indicate that MinerU is used. \\
      \midrule
      Model & GPT-4o-mini & Response generation & OpenAI~\cite{openai2024gpt4omini} & Proprietary hosted model/service accessed via API. Usage is subject to the applicable OpenAI service terms and policies; no model weights are redistributed in this work. \\
      Model & Qwen3-Embedding-4B & Text embedding & Zhang et al.~\cite{zhang2025qwen3} & Apache-2.0. \\
      Model & PE-Core-G14-448 & Image embedding & Bolya et al.~\cite{bolya2025perception} & Apache-2.0. \\
      Model & colnomic-embed-multimodal-3b & Page-level embedding & Nomic AI~\cite{nomic2024colnomic} & Qwen Research License Agreement. It permits non-commercial research/evaluation use; commercial use requires a separate license. Redistributions must retain the required notices, and products that use it to build or improve an AI model should prominently display the required Qwen attribution. \\
      Model & bge-reranker-v2-m3 & Text reranking & BAAI~\cite{baai2024bgererankerv2m3} & Apache-2.0. \\
      Model & Qwen3-VL-32B-Instruct & KG extraction; response generation & Bai et al.~\cite{bai2025qwen3} & Apache-2.0. \\
      Model & Qwen3-32B & LLM-based answer scoring & Qwen Team~\cite{qwen2025qwen332b} & Apache-2.0. \\
      \bottomrule
    \end{tabular}
  }
\end{table*}

\section{Limitations}
\label{app:limitations}

Despite the substantial advancements achieved, BayesRAG is currently subject to certain limitations that point towards future research directions. First, we observe a \textit{Retrieval-Generation Gap}. While our Bayesian retriever achieves high effective recall (61.7\% @ Top-10 and 76.6\% @ Top-20 in Table~\ref{tab:retrieval_recall}), the downstream generation accuracy is bounded by the current reasoning capabilities of MLLMs when interpreting information-dense visual evidence. We anticipate that as generator models evolve, they will better utilize the high-fidelity evidence provided by our framework.
Second, as our methodology is specialized for \textit{multimodal} retrieval, its advantages are naturally less pronounced in domains dominated by pure text or massive continuous tables. In such contexts, the standard chunking strategy often fragments tabular continuity. This remains a common challenge inherent to most RAG pipelines. BayesRAG prioritizes cross-modal consistency, which is most effective when documents contain rich interplay between text and images.

%%%%%%%%%%%%%%%%%%%%%%%%%%%%%%%%%%%%%%%%%%%%%%%%%%%%%%%%%%%%

% \newpage
% \input{checklist.tex}

\end{document}